\documentclass[letterpaper]{article} 

\ifdefined\aaaianonymous
    \usepackage[submission]{aaai2026}  
    
\else
    \usepackage{aaai2026}              
\fi
\usepackage{multirow}
\usepackage{bbding}
\usepackage{times}  
\usepackage{booktabs} 
\usepackage{helvet}  
\usepackage{courier}  
\usepackage[hyphens]{url}  
\usepackage{graphicx} 
\usepackage{amsmath}
\usepackage{natbib}  
\usepackage{caption} 
\usepackage{amssymb} 
\usepackage[pagebackref,breaklinks,colorlinks,allcolors=iccvblue]{}
\newcommand{\ljxrebu}[1]{{\color{black}#1}}
\usepackage{algorithm}
\usepackage{algorithmic}

\usepackage{newfloat}
\usepackage{listings}
\DeclareCaptionStyle{ruled}{labelfont=normalfont,labelsep=colon,strut=off} 
\floatstyle{ruled}
\newfloat{listing}{tb}{lst}{}
\floatname{listing}{Listing}

\ifdefined\aaaianonymous
    \title{ TubeRMC: \underline{Tube}-conditioned \underline{R}econstruction with  \underline{M}utual \underline{C}onstraints for Weakly-supervised Spatio-Temporal Video Grounding}
\else
    \title{ TubeRMC: \underline{Tube}-conditioned \underline{R}econstruction with  \underline{M}utual \underline{C}onstraints for Weakly-supervised Spatio-Temporal Video Grounding}
\fi

\author {
    Jinxuan Li\textsuperscript{\rm 1},
    Yi Zhang\textsuperscript{\rm 1},
    Jian-Fang Hu\textsuperscript{\rm 1,2,3}\thanks{Corresponding author}
    Chaolei Tan\textsuperscript{\rm 4}
    Tianming Liang\textsuperscript{\rm 1}
    Beihao Xia\textsuperscript{\rm 5}
}
\affiliations {
    \textsuperscript{\rm 1}Sun Yat-sen University,
    \textsuperscript{\rm 2}
    Guangdong Province Key Laboratory of Information Security Technology, China,\\
        \textsuperscript{\rm 3}Key Laboratory of Machine Intelligence and Advanced Computing, Ministry of Education, China,\\
    \textsuperscript{\rm 4}The Hong Kong University of Science and Technology,
        \textsuperscript{\rm 5}Huazhong University of Science and Technology,
        \\
    \{lijx267,zhangy2799,liangtm9\}@mail2.sysu.edu.cn, hujf5@mail.sysu.edu.cn, ctanak@connect.ust.hk, xbh\_hust@hust.edu.cn
}

\usepackage{bibentry}

\begin{document}

\maketitle

\begin{abstract}
Spatio-Temporal Video Grounding (STVG) aims to localize a spatio-temporal tube that corresponds to a given language query in an untrimmed video. This is a challenging task since it involves complex vision-language understanding and spatiotemporal reasoning. To eliminate reliance on fine-grained annotations like bounding boxes or temporal stamps, recent works have explored  weakly-supervised setting in STVG. However, they typically 
follow a simple late-fusion manner, which generates tubes independent of the text description, often resulting in failed target identification and inconsistent target tracking. 
To address this limitation, we propose a Tube-conditioned Reconstruction with Mutual Constraints (\textbf{TubeRMC})  framework that generates text-conditioned candidate tubes with pre-trained visual grounding models and further refine them via tube-conditioned reconstruction with spatio-temporal constraints. Specifically, we design three reconstruction strategies  from temporal, spatial, and spatio-temporal perspectives to comprehensively capture rich tube-text correspondences. Each strategy is equipped with  a  Tube-conditioned Reconstructor, utilizing spatio-temporal tubes as condition to reconstruct the key clues in the query.  We further introduce mutual constraints between spatial and temporal proposals to enhance their quality for reconstruction. TubeRMC outperforms existing methods on two public benchmarks VidSTG and HCSTVG. Further visualization shows that TubeRMC effectively mitigates both target identification errors and inconsistent tracking.
\end{abstract}
\begin{links}
\link{Extended version}{https://arxiv.org/abs/2511.10241}
\end{links}
\ifdefined\aaaianonymous
\else

\section{Introduction}
\label{sec:intro}

\begin{figure} [t!]
	\centering
	\includegraphics[width=\linewidth,height=0.32\textheight]{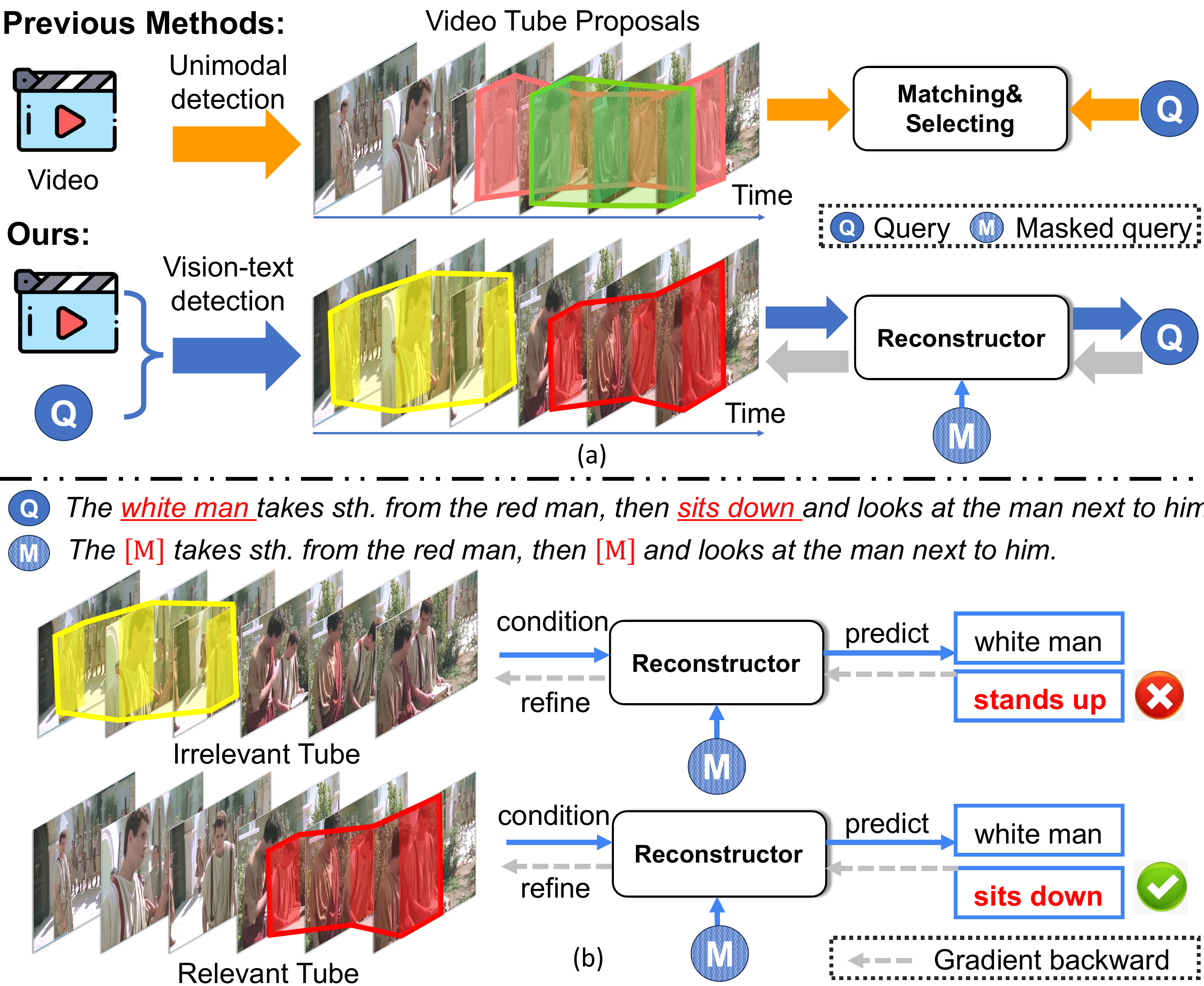}
	\caption{\ljxrebu{\textbf{(a)} Comparison with previous WSTVG methods.} 
   \textbf{(b)} Illustration of tube-conditioned reconstruction. The tube that matches the query descriptions can correctly reconstruct  masked phrases. Moreover, the process can  refine tubes predicted in detection. {\texttt{[M]}} means the masked phrase. }
	\label{fig:introduction}
\end{figure}

STVG aims to predict a spatio-temporal tube (i.e., a sequence of bounding boxes within a specified time interval) corresponding to the event described by a language query in an untrimmed video. 
While existing methods \cite{TubeDETR,stvg2} have achieved remarkable progress, they rely heavily on expensive tube-text annotations for supervised learning.  
To address this limitation, recent methods \cite{li2023winner,eccv2024} have explored weakly-supervised STVG (WSTVG), which relies solely on video-text pairs without requiring bounding boxes or temporal annotations during training.
These works typically follow a simple late-fusion manner, which employs a unimodal detector like Faster-RCNN \cite{faster-rcnn} to produce  tube proposals and then  matches them with the input query, as shown in Figure \ref{fig:introduction}(a). A main limitation of them is the  tubes are generated independent of the text description, often resulting in failed target identification.\cite{li2025image}


To overcome the limitation, we  propose to introduce pre-trained visual grounding models, which can capture text-conditioned object localizations and provide more reliable spatial grounding results. A straightforward way to  apply these models is concatenating frame-wise results to form spatio-temporal tubes.
However, this is ineffective for WSTVG since the object identification could be  inconsistent across frames and  temporal boundaries often fail to capture target event due to the lack of spatiotemporal understanding. 

To fully unleash the potential of visual grounding models in WSTVG, we propose a novel framework \textbf{TubeRMC} that employs \textit{tube-conditioned reconstruction}  to enhance tube-text correspondence learning and  event understanding, while utilizing \textit{spatio-temporal mutual constraints} to refine both bounding boxes and temporal boundaries.  The tube-conditioned reconstruction explores rich semantic dependencies between visual cues and textual descriptions by reconstructing masked text conditional on tube, learning the tube-text correspondences and achieving event understanding.  
In this way, the model learns to capture the tube-text correspondences by selecting the best tube proposal to reconstruct the text. 
As shown in  Figure \ref{fig:introduction}(b), 
 the tube corresponding to  the target event can reconstruct the key phrases `white man' and `sits down' in the masked sentence.   
More specifically, instead of  employing  only temporal reconstruction like \cite{cao2023iterative,kim2024gaussian}, we propose three reconstruction strategies from temporal, spatial, and spatio-temporal perspectives to capture more comprehensive tube-text correspondences.  This is achieved by reconstructing key phrases from the text, conditioned on tube features generated under the guidance of 1D, 2D, and 3D Gaussian attention maps. Each strategy employs a Tube-conditioned Reconstructor, which leverages tube features to reconstruct the masked phrases from visual representations.

\ljxrebu{To further enhance the quality of tube proposals, we introduce mutual constraints consisting of:
(1) a time-to-space constraint that ensures motion continuity for objects within the same scene, and
(2) a space-to-time constraint that enforces temporal boundaries to align with high-confidence frames.
These complementary constraints jointly improve object consistency and temporal prediction accuracy.}

Extensive experiments on four public STVG benchmarks show significant improvement of TubeRMC over state-of-the-art (SOTA) methods. For example, in the HCSTVG-v1 dataset, our TubeRMC surpasses the previous SOTA method, VCMA \cite{eccv2024}, by 4.74\% in the m\_vIoU metric. Moreover, compared to the baseline that directly using MDETR, TubeRMC demonstrates significant performance advantages. For instance, on the HCSTVG-v2 dataset, TubeRMC improves the baseline model by over 8\% in terms of m\_vIoU. These results highlight the effectiveness of our tube-conditioned reconstruction with mutual constraints framework.  
In summary, our contributions are: 

\begin{itemize}
\item[$\bullet$] We propose a novel TubeRMC framework to learn fine-grained tube-text alignment without requiring annotations of bounding boxes or temporal intervals.

\ljxrebu{\item[$\bullet$] We propose tube-conditioned reconstruction to comprehensively capture tube-text correspondences from 1D, 2D, 3D (temporal, spatial, spatio-temporal) perspectives, with  mutual constraints to enhance proposal quality.}


\item [$\bullet$]Our approach outperforms previous state-of-the-art methods on VidSTG  and HCSTVG benchmarks.  
\end{itemize}

\begin{figure*} [t!]
	\centering
	\includegraphics[width=\linewidth,height=0.34\textheight]{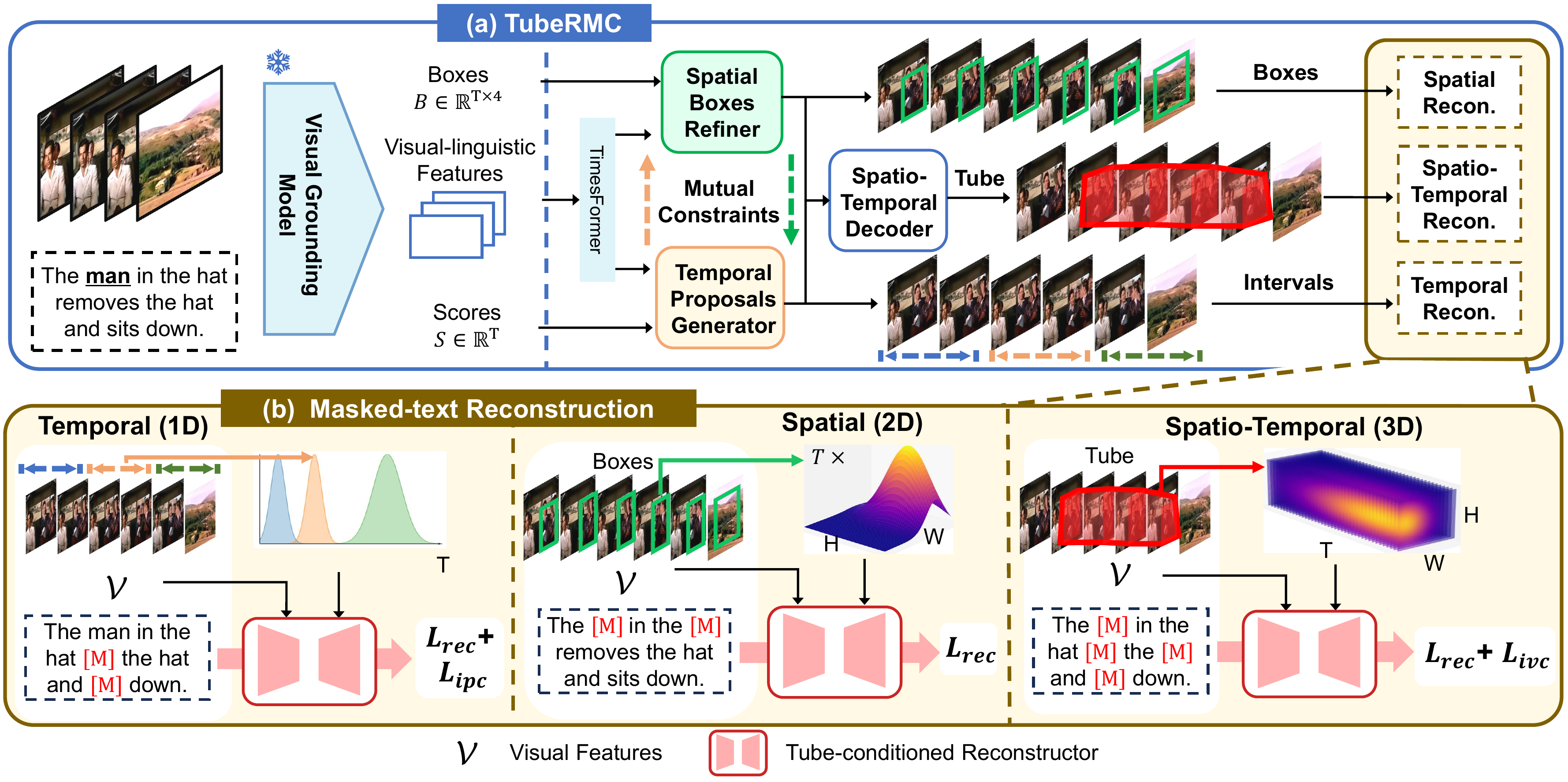}
	\caption{\textbf{(a) TubeRMC Overview.} It selects the most relevant bounding box for the subject token in each frame while extracting image-level visual-linguistic features (left), and generates temporal intervals (1D), spatial bounding boxes (2D), and spatio-temporal tube proposals (3D) for reconstruction (right).
\textbf{(b) Tube-conditioned Reconstruction Learning} masks out phrases containing dynamic, static, or holistic event information in the sentence, then reconstructs the masked phrases condition on 1D, 2D, and 3D proposals. It further introduces mutual constraints between 1D and 2D proposals, employing time-to-space and space-to-time constraints to enhance spatio-temporal consistency in proposal generation.}
	\label{fig:framework} 
\end{figure*}

\section{Related Work}

\noindent \textbf{Fully-supervised STVG.}
Spatio-Temporal Video Grounding focuses on identifying the target object both spatially and temporally based on a language query. Earlier methods \cite{where-dose-it-exist, obj-aware, HCSTVG_paper} use  detectors like Faster R-CNN to detect objects in each video frame and then ground the temporal locations based on the detection features. STGRN \cite{where-dose-it-exist} and OMRN \cite{obj-aware} capture object relations through spatio-temporal graphs and multi-branch relation networks, respectively. Recent works  \cite{TubeDETR,tan2021augmented,STCAT,Lin_2023_CVPR,videogroundingDINO,liang2024costa,stvg1,stvg2,liang2025referdino,liang2025long} have developed frameworks based on powerful pre-trained DETR-based \cite{detr} vision-language models with additional spatio-temporal interactions. These methods depend heavily on manual annotations, which is quite costly. 

\noindent \textbf{Weakly-supervised STVG.}
Many researchers propose to address Spatio-Temporal Video Grounding with less annotations  \cite{wsstg,adws,li2023winner,visctx,eccv2024} and develop weakly-supervised spatio-temporal grounding models. ADWS \cite{adws} provides a framework with mutually-guided spatio-temporal Multiple-Instance Learning to match each query with specific spatial regions in video frames. WINNER \cite{li2023winner} proposes a hierarchical video-language decomposition-alignment structure for multi-modal matching. VCMA \cite{eccv2024} uses the variational Expectation-Maximization algorithm to rebuild visual relationships between entities and achieve more accurate cross-modal alignment. 
However, they typically follow a detect-then-match process, where spatial proposals are generated by a unimodal detector like Faster-RCNN \cite{faster-rcnn}, tracked into tubes, and matched with text-based similarity. This ignores the unique correspondences between video regions and sentence components, which is crucial for WSTVG, limiting their performances. \ljxrebu{\cite{garg2025stpro} directly employs G-DINO's \cite{groundingdino} spatial detections to construct tubelets via tracking and applies progressive learning for temporal modeling. However, its performance is limited by inherited noise from G-DINO, as it cannot refine G-DINO's detection results.}

\noindent \textbf{Reconstruction for Weakly-supervised Video Grounding.} Reconstruction-based methods  \cite{lin2020weakly,zheng2022cnm,cao2023iterative,kim2024gaussian,weakCVPR2022} have been extensively studied for weakly-supervised video temporal grounding (WVTG) \cite{wtg2,wtg3,wtg4,wtg6,wtg8,wtg9,wtg10,wtg11,wtg12,wtg13,wtg14}.  
It assumes that temporal proposals matched well with the input text can reconstruct a sentence query from a randomly masked sentence query. \cite{zheng2022cnm} proposes a learnable Gaussian attention mask to generate proposals. It further introduces a mask-conditioned mechanism that restricts attention to the frames specified by the mask. \cite{kim2024gaussian} extends basic Gaussian proposals to Gaussian mixture models. However, these WVTG methods only focus on temporal masks and ignore the correspondences between spatial information and text, which is crucial for STVG task. In this work, we propose a novel Tube-conditioned Reconstructor that can simultaneously handle both temporal and spatial attention masks.

\section{Methodology}


\subsection{Model Overview}
In Figure \ref{fig:framework}, we present our TubeRMC. 
We first introduce the \textbf{model architecture} of TubeRMC. It first employs a pre-trained visual grounding model to extract image-text correspondences and spatial grounding results per frame, and then performs  cross-frame modeling to capture spatio-temporal context and generate diverse proposals for reconstruction.
To learn fine-grained tube-text alignment without spatial or temporal annotations, we propose \textbf{Tube-conditioned Reconstruction Learning}. This learning scheme primarily incorporates three novel reconstruction strategies that rebuild masked phrases from temporal, spatial, and spatio-temporal perspectives respectively, comprehensively capturing spatio-temporal correspondences between visual cues and query descriptions. Furthermore, we introduce mutual constraints to enhance proposal quality for the reconstruction task.

\subsection{Model Architecture}
\label{sec:mdetr}
\noindent \textbf{Static Cross-modal Extraction.} We employ a pre-trained  visual grounding model to extract frame-wise cross-modal representations and grounding results including bounding boxes and confidence scores. Recent visual grounding models MDETR\cite{mdetr}, G-DINO adopt the DETR-based \cite{detr} architecture, including an image backbone, a text encoder, and a transformer encoder-decoder with a box head and a confidence score head.  These models take an image and a text query as input and output several  bounding boxes with confidence scores for each query token. For \underline{each image}, we rank all boxes by scores for the subject token and pick the highest-scoring box as the prediction for the whole query.  
We then concatenate the selected box in each frame and form a bounding box tube \(B \in \mathbb{R}^{T \times 4}\) with confidence score vector \(S \in \mathbb{R}^{T \times 1}\) where $T$ is the number of frames. 
\ljxrebu{
For target-invisable frames, MDETR tends to produce low-confidence regions, owing to its ability to align visual regions with text. These confidence scores can be utlized in  our framework to filter out non-target regions.}



\ljxrebu{\noindent \textbf{Spatio-Temporal Modeling.} The cross-modal features 
obtained from the grounding model are fed into the cross-modal TimesFormer \cite{TimesFormer,Lin_2023_CVPR}, obtaining the cross-modal features \( F_t \in \mathbb{R}^{T \times (H\times W+L)\times d}\) and the global frame features \( F_g \in \mathbb{R}^{T\times d} \). \( H \) and \( W \) denote the  height and width of video frames, respectively, while \( L \) represents the length of input query. These features are then fed into Spatial Boxes Refiner and Temporal Proposals Generator to refine spatial grounding results and generate temporal proposals, respectively. In the Spatial Boxes Refiner, the object features (queries) of selected boxes from MDETR are processed with $F_g$ via a cross-modal attention mechanism to capture inter-frame contextual relationships. 
The output queries  are passed through a regression head to predict offsets that refine MDETR's outputs, yielding spatial proposal boxes and refined confidence scores.
In Temporal Proposals Generator, we use $K$ learnable queries in a temporal decoder to model temporal context. The architecture of decoder is similar to that of the spatial refiner, with key difference being that both \( F_t \) and \( F_g \) are used simultaneously in the cross-attention mechanism. 
The output embeddings are then passed through a fully-connected layer to predict  temporal  proposals.   

\begin{figure}[t]
  \centering
\includegraphics[width=0.8\linewidth,height=0.144\textheight]{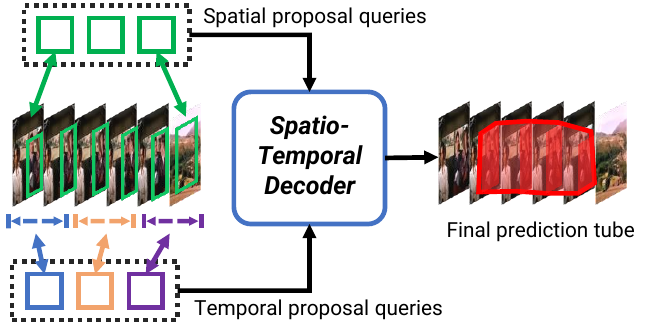}
  \caption{The effect of the Spatio-Temporal Decoder.} 
  \label{fig:STIC}
\end{figure}
The above pipeline provides basic spatial and temporal contextual modeling separately. However, since WSTVG task requires global video-text correspondences rather than individual spatial or temporal predictions, we further propose a Spatio-Temporal Decoder to integrate both spatial and temporal  information and perform event-level predictions as shown in Figure \ref{fig:STIC}.  
The spatial or temporal visual context in each proposal,  embedded in spatial or temporal queries, is aggregated by a spatio-temporal query for global correspondence learning. 
We then use a spatio-temporal tube head to obtain final prediction. 
The spatial queries are employed for spatial localization, while the spatio-temporal query generates temporal boundaries.} 




\subsection{Tube-conditioned Reconstruction Learning} Unlike previous reconstruction-based methods  \cite{zheng2022cnm,cao2023iterative,kim2024gaussian} which only focus on temporal grounding, we develop a Tube-conditioned Reconstructor (TR) that simultaneously take temporal and  spatial masks as inputs, facilitating the model learning rich spatio-temporal visual-linguistic correspondences. 
Based on TR, we introduce three reconstruction strategies (Temporal, Spatial, and Spatio-Temporal) to capture tube-text correspondences from 1D, 2D, and 3D perspectives.
\begin{figure}[t]
  \centering
  \includegraphics[width=0.9\linewidth,height=0.32\textheight]{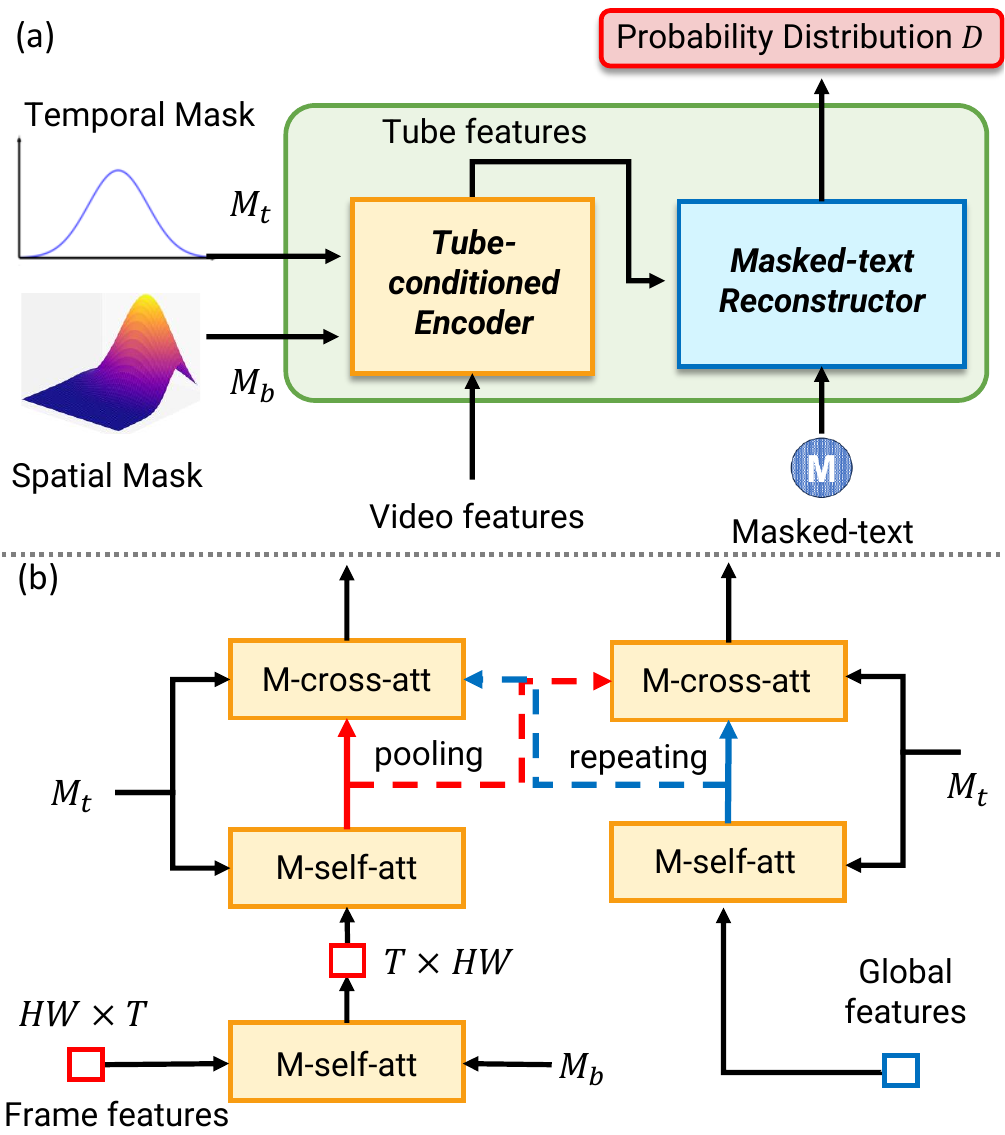}
  \caption{Architecture of Tube-conditioned Reconstructor: (a) Overview; (b) Tube-conditioned Encoder layer design.} 
  \label{fig:TR}
\end{figure}

\noindent \textbf{Representing Spatio-Temporal Masks as Gaussians.} 
To enable gradient back-propagation in TR, we transform temporal intervals, spatial boxes, and spatio-temporal tubes into Gaussian distributions. For temporal intervals in Temporal Proposals Generator, we use a 1D Gaussian function to get temporal attention mask \( M_t \in \mathbb{R}^{T} \). For spatial boxes in Spatial Boxes Refiner, we apply 2D Gaussian function for each frame, treating the center coordinates as the mean and the width as the standard deviation to produce \( M_b \in \mathbb{R}^{HW} \). For the spatio-temporal tube, we employ 3D Gaussian function, treating the tube center coordinates as the distribution center and the tube's length, width, and height as the standard deviations, formulating masks of size \({T\times HW}\).


\noindent \textbf{Tube-conditioned Reconstructor.} As shown in Figure \ref{fig:TR}(a), TR consists of a  Tube-conditioned Encoder and a Masked-text Reconstructor. We feed frame features and global features from TimesFormer 
into the encoder to model spatio-temporal dependencies under the guidance of spatial and temporal masks. Specifically, we define Mask-Attention as:
\begin{align}
M\text{-}att(Q,K,V,M) = (\text{softmax}(\frac{QK^\mathrm{T}}{\sqrt{d}})\bigotimes M)V, 
\end{align}
where $M$ is the temporal or spatial Gaussian mask and $\bigotimes$ means multiplying by each row. As illustrated in Figure \ref{fig:TR}(b), the Tube-conditioned Encoder consists of a local branch (left) that focuses on local temporal and spatial context, and a global branch (right) that complements the understanding of global temporal information. Following this architecture, the encoder focuses on the visual regions corresponding to the masks, reducing responses to irrelevant areas.
Finally, we apply spatial pooling to the visual features output from the final layer, resulting in spatio-temporal tube features $F_T \in \mathbb{R}^{T\times d}$. Different combinations of temporal and spatial masks enable the encoder to focus on different regions of the tube in video, producing different $F_T$. 

The tube features $F_T$ are then passed through the Masked-text Reconstructor. 
In the reconstructor, we first replace some word tokens
in the original query with a special \texttt{[MASK]} token. The masked-text features are then fed into a transformer-decoder-like reconstructor to acquire the cross-modal semantic representations. Next, we apply a FFN with a softmax function to obtain the probability distribution $D$ for the masked words in the vocabulary.


\noindent\textbf{Reconstruction Strategies.} 
To capture comprehensive  spatio-temporal correspondences between visual cues and linguistic descriptions, we construct three types of reconstruction strategies including Temporal, Spatial, and Spatio-Temporal Reconstruction. For \underline{Temporal Reconstruction}, predicates and their directly related nouns are selectively masked, which mainly includes motion-aware temporal information. The positive proposals are then transformed into 1D Gaussian temporal attention masks, respectively. 
For \underline{Spatial Reconstruction}, we mask subject nouns with adjectives in the query and use the 2D spatial Gaussian Masks \(M_b \) to reconstruct the masked texts. This helps the model focus on the correspondence between object appearance information in both the video and the text. 
For \underline{Spatio-Temporal Reconstruction}, we utilize 3D  Gaussian masks for TR and randomly mask a set of words across sentence, with a higher probability of masking verbs, nouns, and adjectives. This enables the model to effectively capture spatio-temporally coupled correspondences.

\noindent \textbf{Reconstruction Loss.} We use cross-entropy loss $L_c(\cdot)$  \cite{lin2020weakly,kim2024gaussian} to measure the difference between the reconstructed query and the ground truth query. The reconstruction loss is formulated as: 
\begin{align}
\setlength{\abovedisplayskip}{.2em}\setlength{\belowdisplayskip}{.2em}
L_{rec}= L_c(D_{s}) + L_c(D_{t})+ L_{c}(D_{g}),
\end{align}
where   $D_{s}, D_{t},$ and $ D_{g} $ represent the predicted distribution from Spatial, Temporal and Spatio-Temporal Reconstruction, respectively. 

\ljxrebu{Furthermore, to enable temporal visual-linguistic understanding without temporal annotations, we introduce a set of negative samples and design an inter-proposal contrastive loss for Temporal Reconstruction. For the construction of negative proposals, please refer to the next subsection. Since negative samples contain a significant amount of visual information unrelated to the query, their corresponding reconstruction loss should be higher than positive samples. Based on this, the inter-proposal contrastive loss $L_{ipc}$ is defined as:
\begin{align}
\setlength{\abovedisplayskip}{.2em}\setlength{\belowdisplayskip}{.2em}
L_{ipc}= \sum_{\substack{i=1, i\neq k^*}}^K\mathrm{max}\bigl(L_c({D}_{t}^{(k^*)})-L_c({D}_{t}^i)+\beta_1,0\bigr) \\
+\sum_{\substack{i=1}}^K\mathrm{max}\bigl(L_c({D}_{t}^{(k^*)})-L_c({D}_{tn}^i)+\beta_2,0\bigr),\notag
\end{align}
where $k^*$ denotes the index of the positive proposal that minimizes the cross-entropy loss and $D_{tn}^i$ means the predicted distribution for $i$-th negative temporal proposal.

Additionally, we generate two negative samples for Spatio-Temporal Reconstruction by perturbing the temporal prediction following \cite{zheng2022cnm}. The easy negative sample uses an inverted temporal mask of the original tube, while the hard negative sample assigns uniform mask (all values set to 1) across the temporal dimension.
The intra-video contrastive loss  can be expressed as:}
\begin{align}
L_{ivc}=
\mathrm{max}\bigl(L_c({D}_{g})-L_c({D}_{ghn})+\beta_3,0\bigr)+ \nonumber\\
\mathrm{max}\bigl(L_c({D}_{g})-L_c({D}_{gen})+\beta_4,0\bigr),  
\end{align} where \( D_{ghn} \), \( D_{gen} \) represent the reconstruction distributions of the hard and easy negative samples in Spatio-Temporal Reconstruction, respectively. The hyper-parameters $\beta_1$ to $\beta_4$ control the margins and $\beta_2 > \beta_1$, $\beta_4 > \beta_3$. 

\ljxrebu{\noindent \textbf{Mutual  Constraints Learning.} 
To improve both object consistency and temporal predictions quality for reconstruction, we propose  mutual constraints learning. 
The space-to-Time constraint leverages spatial confidence scores to guide the generation of more precise temporal proposals. 
We first use scores \(\hat{S}\) from Spatial Boxes Refiner to generate positive proposals.  Specifically, the initial proposals are generated by selecting the top \( K \) scoring frames as temporal midpoints  with a predefined width \(w_i\). The Proposals Generator then predicts midpoint and width offsets,  which are applied to yield the final positive proposals \(\in \mathbb{R}^{K\times 2} \). Similarly, we construct  negative temporal proposals  using frames with the \(K\)-lowest scores. 
On this basis, the space-to-time constraint loss is defined to minimize overlaps both between different positive proposals and between positive and negative proposals. 
Furthermore, we propose the time-to-space constraint to make sure that objects maintain spatial continuity between adjacent frames within each scene. This is achieved by applying a loss that penalizes predicted boxes in adjacent frames with an IoU below a threshold within each temporal proposal.} 

\subsection{Model Training and Inference}

We train our model for weakly-supervised Spatio-Temporal Video Grounding in an end-to-end manner by minimizing the following loss function:
\begin{align}
\setlength{\abovedisplayskip}{-.8em}\setlength{\belowdisplayskip}{1em}
L_{total}=L_{rec} + L_{ipc} + L_{ivc}+L_{mc},
\end{align}
where $L_{mc}$ denotes mutual constraints loss, introduced to guide the refinement of both spatial and temporal proposals. 
During inference, we use the spatio-temporal tube output by Spatio-Temporal Decoder as the final grounding results.

\begin{table*}[ht]
\centering
  \footnotesize
\setlength{\tabcolsep}{3pt}
\begin{tabular}{l|ccc|ccc|ccc}
\toprule
\multirow{2}{*}{Methods}   &\multicolumn{3}{c|}{HCSTVG-v1} &\multicolumn{3}{c|}{VidSTG Declarative} & \multicolumn{3}{c}{VidSTG Interrogative}  \\ & m\_vIoU & vIoU@0.3 & vIoU@0.5 & m\_vIoU & vIoU@0.3 & vIoU@0.5 & m\_vIoU & vIoU@0.3 & vIoU@0.5  \\ \hline
\textit{Two stages pipelines} \\
\hline
GroundeR
+LCNet &4.17 &3.28 &1.05& 7.85 &7.96 &3.02 &6.43 &6.58& 2.92\\
MATN+LCNet &4.41& 3.53& 1.12 &8.16 &8.03 &3.59 &6.97 &6.64 &3.05\\
GroundeR+CPL & 5.23& 4.18& 1.25 &8.28& 8.35 &3.68 &7.16& 7.28& 3.23\\
RAIR+CPL 
&6.88 &4.87 &1.36& 8.67& 8.72& 4.01& 7.68 &7.71 &3.58 \\
\hline
 \textit{One stage pipelines} \\
\hline
WSSTG 
& 6.52 & 4.54 & 1.27 & 8.85 & 8.52 & 3.87 & 7.12 & 6.87 & 2.96  \\
ADWS 
& 8.20 & 4.48 & 0.78 & 8.96 & 7.86 & 3.10 & 8.57 & 8.64 & 2.88  \\
Vis-Ctx 
&9.76 & 6.81 & 1.03  & 9.34 & 7.32 & 3.34 & 8.69 & 7.18 & 2.91   \\
WINNER 
& 14.20 & 17.24 & 6.12 & 11.62 & 14.12 & 7.64 & 10.23 & 11.96 & 5.46 \\
VCMA 
&14.64&18.60&5.75& 14.45 & 18.57 & 8.76 & 13.25 & 16.74 & \textbf{7.66} \\ 
\hline
MDETR-Zero&12.26&18.10&5.03&11.13&14.41&6.06&6.52&7.80 & 2.96\\ 
MDETR+CPL   & 15.27 & 17.93&3.61  &13.02&18.89&3.45 & 8.64 &7.94 & 3.49\\
\hline
TubeRMC (Ours) & \textbf{19.38}  & \textbf{23.88} &  \textbf{6.75}& \textbf{15.93}& \textbf{25.16} &\textbf{9.09}& \textbf{13.47} & \textbf{18.79}& 7.64 \\ \toprule
\end{tabular}
\caption{Performance comparisons of the state-of-the-art methods on the VidSTG and the HCSTVG-v1 test set (\%). 
}
\label{tab:vidstg}
\end{table*}


\section{Experiments}
\subsection{Experimental Settings}
\noindent \textbf{Datasets.} We perform comprehensive experiments on the HCSTVG \cite{HCSTVG_paper} and VidSTG \cite{where-dose-it-exist} datasets. 
The HCSTVG-v1 consists of 4,500 video-text pairs for training and 1,160 for testing. The HCSTVG-v2 expands on v1 with improved annotation, containing 10,131 pairs for training, 2,000 for validation, and 4,413 for testing. Since the test annotations for HCSTVG-v2 are not publicly available, we report our results on the validation set. The VidSTG dataset includes 99,943 video-text pairs, comprising 44,808 declarative sentences and 55,135 interrogative sentences. The training, validation, and test sets consist of 80,684, 8,956, and 10,303 sentences respectively, along with 5,436, 602, and 732 videos respectively. 

\noindent \textbf{Evaluation Metrics.}
We follow \cite{where-dose-it-exist,li2023winner} and use mean vIoU as our main metric, which is calculated as:
\begin{math} vIoU = \frac{1}{\left|T_u\right|}\sum_{t\in T_i}IoU(\hat{b}_t, b_t), \end{math}
where $T_i$ represents the intersection and $T_u$ represents the union of the time intervals derived from the annotations and the predictions. $\hat{b}_t$ and $b_t$ refer to the predicted and ground truth bounding boxes for the $t$-th frame, respectively. The vIoU score is averaged across all samples to obtain the mean vIoU (m\_vIoU). Additionally, we report vIoU@R, which indicates the percentage of samples with a vIoU score greater than $R$. We also follow \cite{TubeDETR} and report the sIoU and tIoU metrics evaluating the spatial and temporal grounding performances in ablation studies, respectively. They are defined as:\begin{math}
    sIoU=\frac{1}{\left|T_{gt}\right|}\sum_{t\in T_{i}}IoU(\hat{b_t}, b_t), tIoU=\frac{\left|T_i\right|}{\left|T_u\right|},
\end{math}
where $T_{gt}$ indicates the ground truth frame set.

\noindent \textbf{Implementation Details.}
We use pre-trained MDETR with ResNet-101 \cite{resnet} as the image encoder and Roberta-base \cite{roberta} as the text encoder.  
$K$ is set to 4 for HCSTVG and VidSTG. We set the number of transformer layers to 3 in Spatio-Temporal modeling, while TR utilizes a 6-layer architecture. The margin weights $ \beta_1, \beta_2, \beta_3, \beta_4$ are set to  0.5, 0.7, 0.5, 0.7, respectively.    


\begin{table}[htbp]
  \footnotesize
\centering

\begin{tabular}{l|ccc}
\toprule
{Methods} & m\_vIoU & vIoU@0.3 & vIoU@0.5  \\ \hline
MDETR-Zero&12.21&17.15&5.80\\
MDETR+CPL &15.09&24.95&6.50\\
\hline
TubeRMC (Ours)  & \textbf{20.64} & \textbf{26.05} & \textbf{8.15}  \\ \toprule
\end{tabular}
\caption{Performances on HCSTVG-v2 validation set (\%).}
\label{tab:performance_hcstvgv2}
\end{table}
\subsection{Experimental Results}
To verify the effectiveness of TubeRMC, we develop two baseline models based on MDETR and compare the performance of TubeRMC with the baselines and previous WSTVG methods on HCSTVG and VidSTG.

\noindent \textbf{Comparison with previous SOTA.} As shown in Table \ref{tab:vidstg}, TubeRMC surpasses previous SOTA two-stage and one-stage methods by a margin of 12.50\% and 4.74\% in m\_vIoU in HCSTVG-v1, respectively.  Furthermore, we also outperform all methods on VidSTG Declarative and Interrogative. These results strongly validate effectiveness of TubeRMC.

\noindent\textbf{Comparison with  baselines.} MDETR-Zero is a zero-shot baseline that relies solely on MDETR predictions and generates temporal boundaries based on frame-wise confidence scores. It achieves performance comparable to the previous SOTA on HCSTVG-v1 and VidSTG Declarative without extra training. 
Moreover, we apply temporal reconstruction from WVTG to MDETR's predictions and propose a new baseline MDETR+CPL. 
As shown in Table \ref{tab:vidstg}, it surpasses VCMA \cite{eccv2024} by 0.63\% in terns of m\_vIoU on HCSTVG-v1. Although these two baselines achieve certain performance gains, the results remain unsatisfactory since they fail to capture spatio-temporal correspondences. Compared to them, TubeRMC demonstrates a significant performance advantage. For example, Table \ref{tab:performance_hcstvgv2} shows that TubeRMC outperforms MDETR-Zero and MDETR+CPL by 8.43\% and 5.55\% in m\_vIoU on HCSTVG-v2, respectively. Furthermore, due to the significant gap between the pretraining corpus of MDETR and VidSTG Interrogative, the two baselines exhibit suboptimal performance on this dataset. However, TubeRMC effectively mitigates this gap and achieves superior results compared to previous approaches.  This demonstrates the effectiveness of our tube-conditioned reconstruction with mutual constraints design.
\begin{table}[htbp]
\footnotesize

\centering
{
\renewcommand{\arraystretch}{1.0}
  \setlength{\tabcolsep}{1mm}{
\begin{tabular}{c|c|c|c|c}
  \toprule
  Foundation Model &Backbone& m\_vIoU & vIoU@0.3 & vIoU@0.5 \\
  \midrule
Faster-RCNN &ResNet-101 & 15.31 & 19.97& 6.06 \\
  MDETR&ResNet-101 & 19.38 & 23.88& 6.75 \\ 
  
  G-DINO &Swin-T &19.47&24.12&7.03\\
  G-DINO &Swin-B&21.15 &26.81&8.11\\
  \bottomrule
\end{tabular}}}
\caption{Evaluation on varied viusal grounding models.}
\label{tab:ablation_mdetr}
\end{table}

\begin{table}[t]
  \centering
  \footnotesize
  \renewcommand{\arraystretch}{1.0}
  \setlength{\tabcolsep}{0.5mm}{
  \begin{tabular}{c|c|c|c|c|c}
      \toprule
      Spatial & Temporal & Spatio-Temporal & m\_vIoU & vIoU@0.3 & vIoU@0.5 \\
      \midrule
      $\times$ &$\times$ & $\times$ & 14.91 & 14.87 & 2.32 \\
      \checkmark &$\times$& $\times$& 15.24 &14.74 & 3.10 \\
      $\times$& \checkmark &$\times$ &17.59&20.25& 4.74\\
      \checkmark& \checkmark & $\times$&18.12&20.43& 5.00\\
      $\times$& $\times$
      & \checkmark  &17.37&20.78&5.17\\
      \checkmark &\checkmark &\checkmark & \textbf{19.38}  & \textbf{23.88} &  \textbf{6.75}\\
      \bottomrule
    \end{tabular}}
    
      \caption{Evaluation on the reconstruction strategies.}
      \label{tab:ablationTR}
\end{table}

\begin{table}[t]
  \centering
  \footnotesize
  \renewcommand{\arraystretch}{1.0}
  \setlength{\tabcolsep}{0.5mm}{
  \begin{tabular}{c|c|c|c|c|c|c|c}
      \toprule
     ID& s-to-t& t-to-s & m\_vIoU & vIoU@0.3 & vIoU@0.5 &m\_tIoU&m\_sIoU\\
      \midrule
     (a)& & &15.87  & 11.98 & 1.90 &26.69 &59.83\\
      (b)&\checkmark& &17.65 & 20.51 & 5.26 &29.04 &59.95\\
     (c) & & \checkmark & 17.07 &14.06 &2.32 & 27.38 & 60.13\\
      (d)&\checkmark& \checkmark & \textbf{19.38}  & \textbf{23.88} &  \textbf{6.75} & \textbf{30.94} &\textbf{61.67}\\
      \bottomrule
    \end{tabular}}
      \caption{Evaluation on the mutual constraints.}
    \label{tab:ablationMC}
\end{table}

 \begin{figure*} [t!]
	\centering
	\includegraphics[width=0.9\linewidth,height=0.32 \textheight]{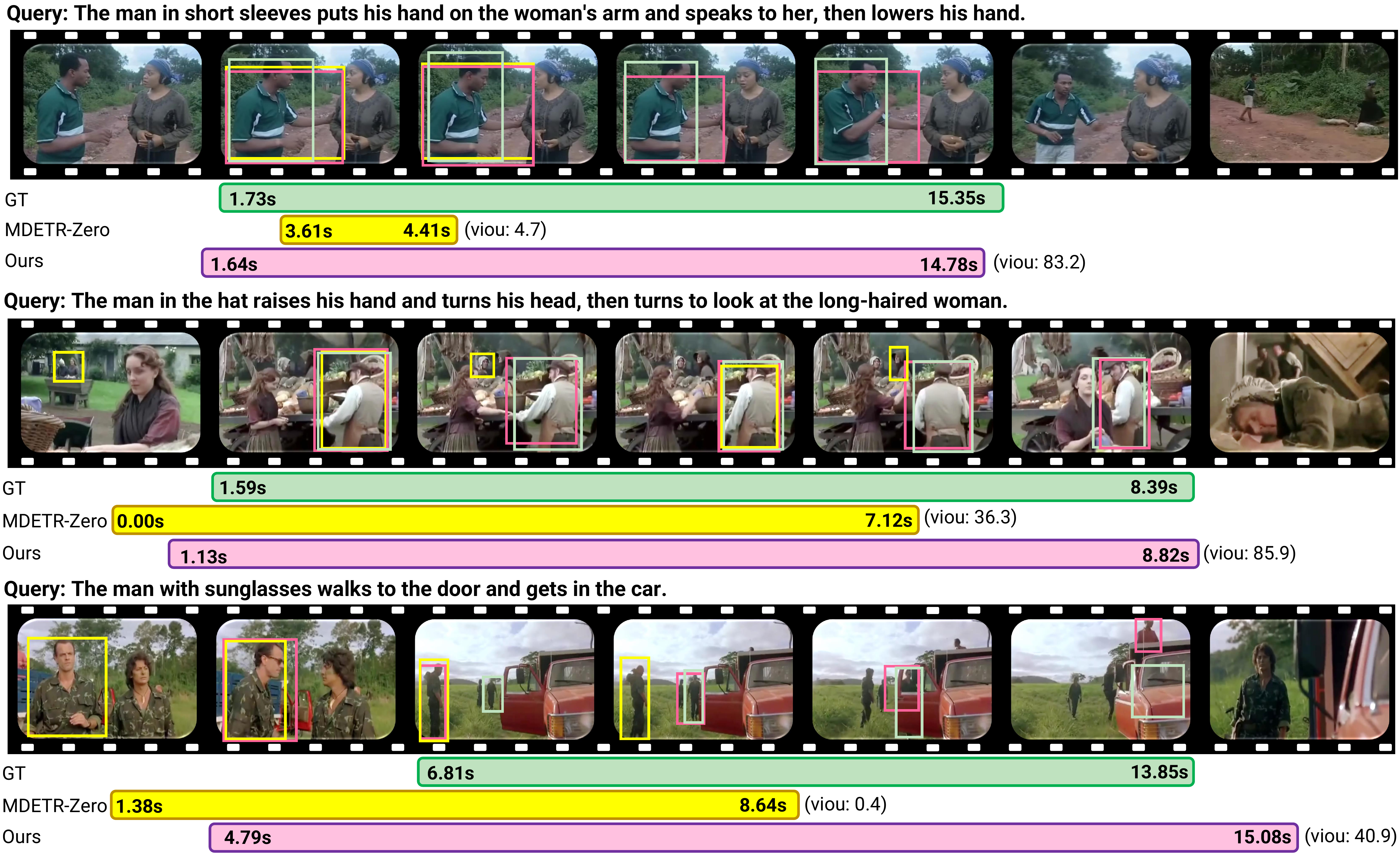}
 \caption{Visualization of spatio-temporal predictions for our method (pink), MDETR-Zero (yellow) and ground truth (green).}
	\label{fig:visualize} 

\end{figure*}

\subsection{Ablation Study}
 In this section, we conduct ablation experiments on HCSTVG-v1 dataset \cite{HCSTVG_paper} to evaluate the effectiveness of some key components in the proposed method.

 \noindent \textbf{Ablation on varied visual grounding models.} To study the effect of foundation models, we replace the visual grounding model MDETR with other models like G-DINO \cite{groundingdino} and ViLBERT (Faster-RCNN) \cite{vilbert}. The results are presented in  Table \ref{tab:ablation_mdetr}.
 We can observe that replacing MDETR with a stronger pre-trained foundation model (e.g., G-DINO-Swin-B) can achieve better performance, improving m\_vIoU from 19.38\% to 21.15\%. To strike a balance between speed and performance, we use MDETR by default.

 \noindent \textbf{Impact of the reconstruction.} Our approach intends to construct three different masked-text reconstruction strategies. Here, we evaluate the effect of our Tube-conditioned Reconstruction by removing the strategies. The experimental results are tabulated in  Table \ref{tab:ablationTR}. As can be observed, either the Spatial or Temporal Reconstruction performs positively to improve system performance. The performance is further improved when both the reconstructions are employed, which means that they can complement well with each other. Specifically, using the Temporal Reconstruction can improve the prediction accuracy by 2.68\% in the term of m\_vIoU, demonstrating the importance of temporal modeling in addressing WSTVG. As expected, combining all the strategies can obtain the best results, enabling the model to capture tube-text correspondences comprehensively.

 \noindent \textbf{Impact of mutual constraints.} We investigate the influences of mutual constraints in  Table \ref{tab:ablationMC}. For experiment (a) and (b), the time-to-space constraint is not included 
 (termed `t-to-s'). In experiments corresponding to row (a) and (c), we exclude the space-to-time constraint (termed `s-to-t').  As shown, our space-to-time constraint boosts the performance of m\_tIoU (from 26.69 to 29.04 and from 27.38 to 30.94) and time-to-space constraint is beneficial for improving spatial grounding results. 
 As expected, using both constraints further improves spatio-temporal prediction performance.

\subsection{Visualization Results}
 We provide several visualization samples predicted by MDETR-Zero baseline and TubeRMC. As shown in Figure \ref{fig:visualize}, MDETR provides unreliable temporal estimates (line 1) and unstable spatial predictions (line 2). In contrast,  TubeRMC captures tube-text correspondences and provides accurate spatio-temporal predictions. This highlights the advantages of tube-conditioned reconstruction in WSTVG learning. Furthermore, we present a challenging case in line 3. The appearance of target `man with sunglasses' is highly ambiguous during GT period. 
 MDETR mistakenly assigns the bounding boxes to another man performing a similar action due to severe viewpoint changes and object occlusions. Although disrupted by MDETR, our TubeRMC still  provides relatively accurate spatio-temporal predictions. We will keep overcoming this issue in our future work. A potential solution is to introduce additional tracking algorithms to guide the model in generating higher-quality tubes. 

\section{Conclusion}
\label{sec:conclusion}

In this work, we propose a novel TubeRMC framework for WSTVG. TubeRMC first deploys a visual grounding model to extract image-text correspondences and then performs cross-frame modeling to capture spatio-
temporal context.  
We propose Tube-conditioned Reconstruction Learning, incorporating three novel reconstruction strategies to capture tube-text correspondences from 1D, 2D, and 3D perspectives. Furthermore, we
introduce mutual constraints to enhance proposal quality.  Our approach  achieves SOTA on VidSTG  and HCSTVG, demonstrating the effectiveness of tube-conditioned reconstruction with  mutual constraints.


\section*{Acknowledgements}
This work was supported partially by the NSFC
 (U22A2095, 6247072922), Guangdong Natural Science Funds Project
 (2023B1515040025), Guangdong NSF for Distinguished Young Scholar
 (2022B15-15020009), and Guangdong Provincial Key Laboratory of Information Security Technology (2023B1212060026). 

\bibliography{aaai2026}

\makeatletter
\fi
\end{document}